\documentclass[conference,table]{IEEEtran}
\IEEEoverridecommandlockouts
\usepackage{bbding}
\usepackage{cite}
\usepackage{amsmath,amssymb,amsfonts}
\usepackage{bm} 
\usepackage{textcomp}
\usepackage{caption}
\usepackage{graphicx}
\usepackage{pgfplots}
\pgfplotsset{compat=1.18}
\usepackage{subcaption}
\usepackage{multirow}
\usepackage{algorithmic}
\usepackage{algorithm}

\usepackage{graphicx}
\usepackage[skip=3pt]{caption} 
\usepackage[font=footnotesize]{caption}
\usepackage{float}
\usepackage{pgfplots}
\usepackage{pgfplots}
\pgfplotsset{compat=1.18}
\usepackage{float}

\usepackage[T1]{fontenc}
\usepackage{booktabs}
\usepackage{multirow}
\usepackage{makecell}
\usepackage{array}
\usepackage{caption}

\pgfplotsset{ 
cycle list={%
{draw=black,mark=star,solid},
{draw=black, mark=square,solid}}}
\usepackage{breqn}
\usepackage{hyperref}
\usepackage{booktabs}

\usepackage{caption}
\usepackage{comment}
\usepackage{authblk}
\usepackage{fancyhdr}
\usepackage{subcaption}
\IEEEoverridecommandlockouts 
\usepackage{fancyhdr}
\fancypagestyle{firstpage}
{
    \fancyhead[L]{\footnotesize © 2026 IEEE.  Personal use of this material is permitted.  Permission from IEEE must be obtained for all other uses, in any current or future media, including reprinting/republishing this material for advertising or promotional purposes, creating new collective works, for resale or redistribution to servers or lists, or reuse of any copyrighted component of this work in other works. This paper is accepted at the Design, Automation and Test in Europe Conference, the European Event for Electronic System Design \& Test (DATE) 2026.}
    \fancyhead[R]{}
}

\begin{document}

\title{HAWX: A \underline{H}ardware-\underline{A}ware Frame\underline{W}ork for Fast and Scalable Appro\underline{X}imation of DNNs}

\author[1]{Samira Nazari}
\author[1]{Mohammad Saeed Almasi}
\author[2,3]{Mahdi Taheri}
\author[1]{Ali Azarpeyvand}
\author[4]{\\Ali Mokhtari}
\author[4]{Ali Mahani}
\author[3]{Christian Herglotz}

 \affil[1]{University of Zanjan, Zanjan, Iran}
  \affil[2]{Brandenburg Technical University, Cottbus, Germany}
 \affil[3]{Tallinn University of Technology, Tallinn, Estonia}
  \affil[4]{Shahid Bahonar University, Kerman, Iran}

\maketitle
\thispagestyle{firstpage}

\begin{abstract}
This work presents \textbf{HAWX}, a hardware-aware scalable exploration framework that employs multi-level sensitivity scoring at different DNN abstraction levels (operator, filter, layer, and model) to guide selective integration of heterogeneous AxC blocks. Supported by predictive models for accuracy, power, and area, HAWX accelerates the evaluation of candidate configurations, achieving over \textbf{23$\times$ speedup} in a layer-level search with two candidate approximate blocks and more than \textbf{(3$\times$10$^{6}$)$\times$ speedup} at the filter-level search only for LeNet-5, while maintaining accuracy comparable to exhaustive search. Experiments across state-of-the-art DNN benchmarks such as VGG-11, ResNet-18, and EfficientNetLite demonstrate that the efficiency benefits of HAWX scale exponentially with network size. The HAWX hardware-aware search algorithm supports both spatial and temporal accelerator architectures, leveraging either off-the-shelf approximate components or customized designs. 
\end{abstract}

\section{Introduction}

Approximate computing (AxC) exploits the error resilience of DNNs, enabling a trade-off between accuracy and efficiency at the hardware level \cite{nourazar2018code, taheri2024adam1, taheri2024exploration}. Innovations such as input-aware arithmetic~\cite{piri2023input}, look-up table sharing multipliers~\cite{guo2024high}, and speculative partial-product units~\cite{hu2024configurable} have broadened the AxC design space, but also introduced scalability issues for design exploration.

To manage this complexity, prior work has explored design space exploration (DSE) and error estimation frameworks, such as DeepAxe \cite{taheri2023deepaxe} and TFApprox \cite{vaverka2020tfapprox}, which enable faster evaluation of approximate designs. Metrics like Architectural Mean Error (AME) \cite{liu2024architectural} and tools like I-NN \cite{ahmadilivani2023special} accelerate error-resilience evaluation. Optimization strategies such as ApproxDARTS (gradient-based) \cite{pinos2024approxdarts}, MAX-DNN (greedy search) \cite{leon2025max}, and reinforcement learning-based DSE frameworks \cite{saeedi2023design} have also been proposed, reducing evaluations from astronomical scales to thousands. However, these methods still demand substantial exploration time and typically overlook heterogeneous approximation or fine-grained search, especially at the filter and operation levels. Addressing speed, scalability, and granularity together remains an open challenge that is addressed in this work.

Moreover, few approaches explicitly account for the diversity of DNN hardware accelerators. Spatial architectures like dataflow accelerators rely on massive parallelism and data reuse \cite{blott2018finn}, while temporal architectures, such as FPGA-based soft processors (e.g., PERUN), depend on flexible time-multiplexed execution \cite{al2016fgpu}. Efficient AxC integration must consider these architectural differences, yet most existing frameworks are hardware-agnostic or target only a narrow class of architectures.

To address these limitations, we propose HAWX, a hardware-aware approximation framework for scalable DSE across heterogeneous accelerators. The main contributions are:
\begin{itemize}
    \item A scalable hardware-aware fully automated framework for rapidly identifying high-quality AxC configurations, supported by a comprehensive predictive model for accuracy, power, and area in DSE.
    \item  A quantitative scoring methodology applied at operation, filter, layer, and model levels, for user-provided AxC blocks.
\end{itemize}

The remainder of this paper is structured as follows: Section II presents the proposed methodology; Section III describes the experimental setup and results; Section IV concludes the paper.

\section{Proposed Methodology}
HAWX is a hardware-aware approximation framework designed to automate DSE for DNNs deployed on heterogeneous hardware platforms. The framework takes as input a pretrained DNN with its target dataset, a library of candidate approximate building blocks such as multipliers and adders, a specification of the target hardware platform (spatial or temporal architecture), and a user-defined accuracy threshold. Using this information, HAWX performs a structured process to analyze the model, assign approximation scores, and propose hardware-specific approximation strategies.

\begin{figure*}[ht]
    \centering
    \includegraphics[width=.8\textwidth]{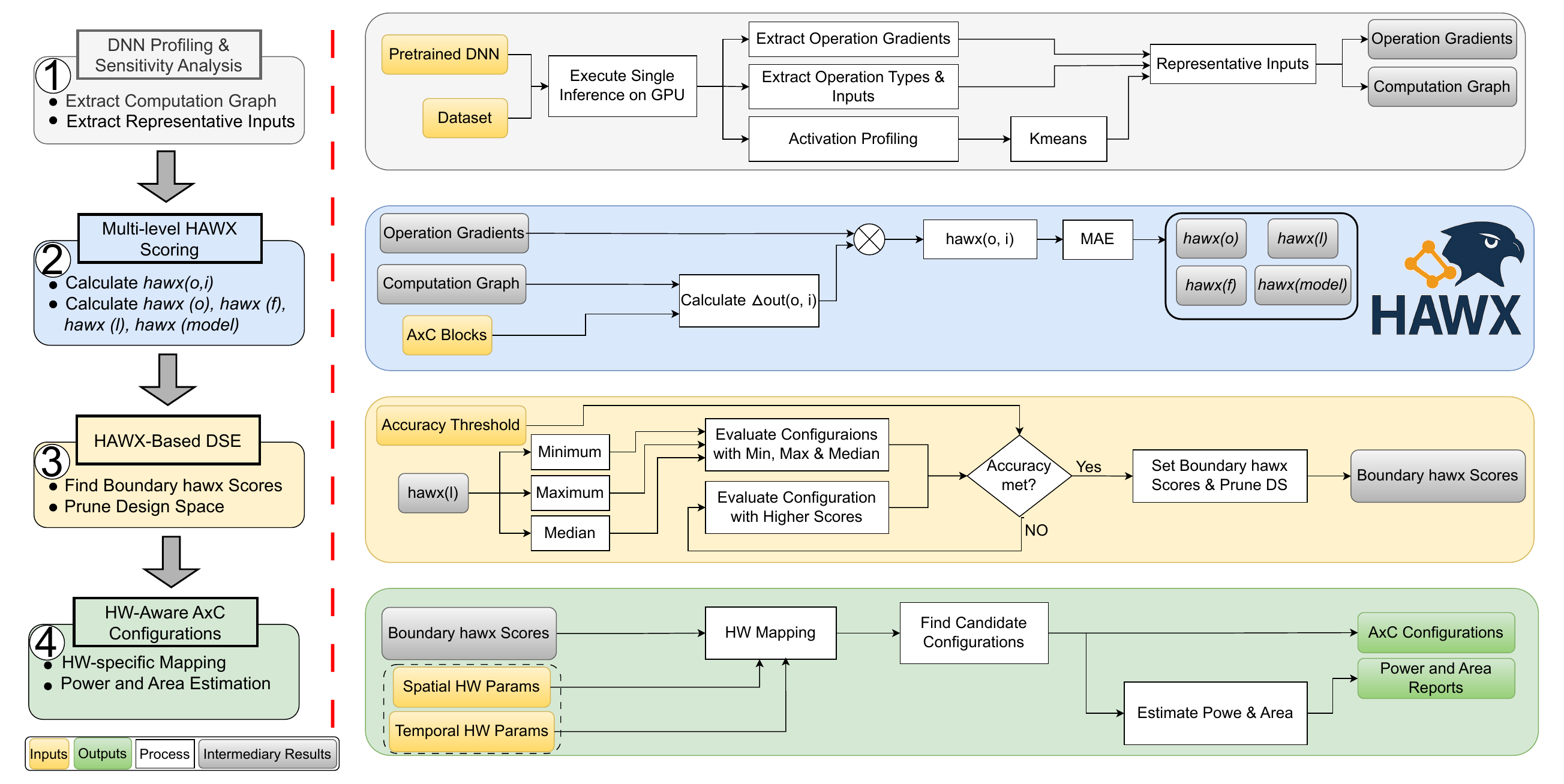}
    \caption{A high-level flow of the proposed methodology}
    \label{fig:hawx_methodology}
\end{figure*}

The first stage of HAWX focuses on profiling the DNN to extract a comprehensive representation of its computation graph (Figure \ref{fig:hawx_methodology}, Step 1). A small subset of input samples is first selected as representative samples by a k-means algorithm to capture the network’s typical data distribution with minimal overhead. The pretrained model is then executed on a GPU using these samples, during which HAWX automatically identifies the types and number of all operations, such as additions and multiplications, along with their inputs. In the same execution pass, the framework computes the gradient of the network loss function with respect to the output of each operation, yielding a fine-grained measure of sensitivity to approximation. This profiling stage thus produces a rich dataset combining structural information about the computation graph with numerical sensitivity metrics, laying the foundation for accurate and hardware-aware approximation analysis while introducing negligible runtime overhead. 

In the second stage, HAWX computes a numerical sensitivity measure, called the $hawx$ score, for every operation in the network (Figure \ref{fig:hawx_methodology}, Step 2). These scores quantify the susceptibility of individual operations to approximation-induced errors. The scoring process is then extended hierarchically to higher levels of abstraction, producing filter-, layer-, and model-level sensitivity metrics. By building this multi-level representation, HAWX creates a scalable foundation for approximation strategies that balance fine-grained precision adjustments (e.g., operator- or filter-level) with coarse-grained design decisions (e.g., layer- or model-level). This enables targeted exploration of large design spaces while maintaining control over both local accuracy impacts and global architectural trade-offs.

The final stages of HAWX leverage the multi-level sensitivity metrics to jointly explore approximation operators (Figure~\ref{fig:hawx_methodology}, Step 3) and hardware mappings (Step 4). Guided by predictive models calibrated for specific hardware platforms, the framework identifies candidate configurations that satisfy the given accuracy constraint. For each viable configuration, HAWX estimates the corresponding power and area costs, producing a set of Pareto-optimal solutions that are guaranteed to meet or exceed the user-defined accuracy threshold. In this way, HAWX transforms a pretrained DNN into a range of hardware-aware approximation configurations through a single inference pass, comprehensive scoring, and hardware cost estimation, streamlining the deployment of approximation techniques across diverse hardware platforms.

\subsection{Approximation Scoring}

The foundation of HAWX is a sensitivity measure that quantifies how vulnerable each operation is to approximation. For an operation $o$ and input $i$, the operation-level score is defined as:

\begin{equation}
\text{hawx}(o, i) = \left\vert \frac{\partial L}{\partial o} \right\vert \times \lvert \Delta out(o, i)\rvert,
\end{equation}

where $\frac{\partial L}{\partial o}$ is the gradient of the loss $L$ with respect to the output of operation $o$, obtained by standard backpropagation, and $\Delta out(o, i)$ is the output error introduced by replacing the exact block with an approximate one:
\begin{equation}
\lvert\Delta out(o, i)\rvert = \lvert out(o, exact)_i - out(o, axc)_i \rvert.
\end{equation}

Since $\Delta out(o, i)$ depends on the specific approximate block employed, this calculation is performed separately for each candidate AxC block, yielding a distinct sensitivity score for every operation–block pair. Thus, the $hawx$ score jointly captures functional sensitivity (via gradients) and numerical deviation (via approximation error), providing a principled metric to rank operations and guide heterogeneous approximation.

The sensitivity of an operation to approximation depends not only on its position in the network but also on the diversity of input data. To ensure robust evaluation, the per-operation $hawx$ scores, $hawx(o,i)$, computed for individual input samples are aggregated into a single representative score per operation. This is achieved by calculating the Mean Absolute Error (MAE) of the per-sample scores across inputs selected by \textit{k-means} clustering:
\begin{equation}
    hawx(o) = \frac{1}{N}\sum_{i=1}^{N}\lvert hawx(o,i) \rvert,
\end{equation}
where $N$ denotes the number of representative input samples.

The selection of these representative inputs aims to capture the diversity of the dataset. To achieve this, activation values from the first network layer are profiled, as this layer is closest to the inputs and thus most sensitive to their variation.
 For each input sample $i$, the activation distribution is modeled as a Gaussian distribution, and the parameters $\mu$ (mean) and $\sigma$ (standard deviation) are computed as:
\begin{equation}
    \mu_i = \frac{1}{M} \sum_{k=1}^{M} a_{i,k},
\end{equation}

\begin{equation}
    \sigma_i = \sqrt{\frac{1}{M} \sum_{k=1}^{M}(a_{i,k} - \mu_i)^{2}},
\end{equation}
where $a_{i,k}$ denotes the activation of the $k$-th neuron in the first layer for input $i$, and $M$ is the total number of activations in that layer.

The pairs $(\mu_i, \sigma_i)$ are then used as feature vectors for k-means clustering, which partitions the dataset into $K$ non-overlapping clusters representing distinct input characteristics. A single representative input is selected from each cluster, and this reduced set of inputs is used to calculate the MAE-based $hawx$ scores for all operations. This process ensures that the computed operation-level sensitivity captures the diversity of the entire dataset while significantly reducing computational overhead, enabling scalable and accurate approximation evaluation.

The operation-level $hawx$ scores are extended to higher architectural levels, i.e., filter, layer, and model, using the MAE aggregation across all operations within each level. For a filter $f$ consisting of $O_f$ operations, the filter-level $hawx$ score is computed as:
\begin{equation}
    hawx(f) = \frac{1}{O_f} \sum_{o \in f} \lvert hawx(o) \rvert.
\end{equation}
The same principle is applied to compute layer-level $hawx(l)$ and model-level $hawx(model)$ scores, where the set of operations corresponds to the operations within the respective layer or the entire network. This approach provides a multi-resolution sensitivity profile of the network, enabling approximation decisions at different granularities.

\subsection{HAWX-based Design Space Exploration}
With $hawx$ scores computed at the operation, filter, layer, and model levels for each approximate block, the next objective is to identify configurations that satisfy the user-defined accuracy threshold. To efficiently search this design space, the process begins by determining boundary score values corresponding to the target accuracy at the layer level. The layer level is chosen because it provides a balance between accuracy, sensitivity, and computational cost. It is more informative than the model level while being faster to evaluate than the operation or filter levels.

To estimate these boundary values, each layer is profiled by evaluating configurations corresponding to the minimum, maximum, and median $hawx$ scores for its approximate blocks. These representative configurations are assembled into three full-network configurations: one using all minimum-score blocks (expected to yield the highest accuracy), one using all maximum-score blocks (expected to yield the lowest accuracy), and one using all median-score blocks (expected to achieve an intermediate accuracy). If these three configurations already identify the score regions that meet or exceed the desired accuracy threshold, the boundaries are assigned directly. Otherwise, additional configurations are explored using a binary search–like strategy, iteratively selecting configurations that step closer to the boundary region based on prior results.

Once the score boundaries are established, approximate block selection for each layer is guided by these values. Only blocks whose $hawx$ scores exceed the identified thresholds are considered, ensuring that the resulting configurations maintain accuracy above the user-specified requirement. This pruning significantly narrows the design space, enabling the subsequent hardware-aware optimization stage to efficiently explore configurations while providing estimated power and area ranges for each candidate.

\textbf{Search Complexity: Exhaustive vs. HAWX}\\ 
Let $B$ be the number of candidate approximate blocks per decision point, and let the network have layers $l = 1, \dots, L$ with $f_l$ filters in layer $l$. In a filter-wise assignment (heterogeneous AxC), each filter independently selects one of $B$ blocks:

\begin{equation}
N_{\text{exh}}^{\text{(filter)}} = \prod_{l=1}^{L} B^{f_l} = B^{\sum_l f_l},
\end{equation}

where $N_{\text{exh}}$ denotes the total number of evaluations required by exhaustive search. Because this number grows exponentially with the total number of filters, exhaustive search quickly becomes intractable in practice. The total runtime of exhaustive search is then:

\begin{equation}
T_{\text{exh}} = N_{\text{exh}} \times t_{\text{eval}},
\end{equation}

where $t_{\text{eval}}$ is the wall-clock time of one evaluation (e.g., fine-tuned inference or testing).  

In contrast, HAWX avoids this exponential blow-up. Its runtime consists of three parts. First, a single profiling pass computes gradients and the model computation graph on representative inputs, taking time $t_{\text{prof}}$, which is significantly smaller than a full evaluation ($t_{\text{prof}} < t_{\text{eval}}$). Second, score computation is performed for all AxC blocks, with total time proportional to the number of blocks $B$, denoted $B \cdot t_{\text{score}}$, where $t_{\text{score}} < 2t_{\text{eval}}$. Finally, only a small and bounded number of full evaluations are required through boundary search, denoted $N_{\text{eval}}^{\text{HAWX}} \cdot t_{\text{eval}}$. Because the design space is analytically pruned by binary search, $N_{\text{eval}}^{\text{HAWX}} \ll N_{\text{exh}}$.

The runtime of HAWX can thus be expressed as:
\begin{equation}
T_{\text{HAWX}} \approx t_{\text{prof}} + B \cdot t_{\text{score}} + N_{\text{eval}}^{\text{HAWX}} \cdot t_{\text{eval}},
\end{equation}

which reduces the search complexity from exponential $O(B^{\sum_l f_l})$ to near-linear in the number of blocks (effectively $O(B)$ with only a small overhead). This makes fine-grained approximation search tractable even for deep networks.

\subsection{Hardware-Aware Approximate Configurations}
After the boundary $hawx$ scores are determined, hardware information is incorporated to map approximate blocks to the target platform. The hardware platform is first identified (Algorithm \ref{alg:hw-aware-axc}, Line 1), such as a spatial (e.g., dataflow)  or temporal accelerator (e.g., PERUN). For each platform, architectural parameters are provided to guide block assignment and estimation of hardware cost.

For spatial accelerators, folding parameters are specified, including the number of Processing Elements (PEs) and SIMD width, which determine the level of parallelism at each layer. The required granularity level (e.g., per-operation, per-filter, or per-layer) is also defined. Based on this granularity, approximate blocks are assigned to operations, filters, or layers that meet the accuracy boundary constraints (Algorithm \ref{alg:hw-aware-axc}, Line 4-6). Power and area are estimated by extracting the count of each approximate block in the configuration and combining this with the base power and area metrics of each block (Algorithm \ref{alg:hw-aware-axc}, Line 7).

For temporal accelerators, hardware descriptions include the number of Compute Vectors (CVs), available logic resources, and interconnect constraints. Because a single CV may execute multiple layers or filters, layers are grouped according to CV availability (Algorithm \ref{alg:hw-aware-axc}, Lines 9-10). Power and area estimations are then computed based on the number of approximate blocks utilized in the grouped execution and their associated hardware costs (Algorithm \ref{alg:hw-aware-axc}, Line 11). In the final stage (Algorithm~\ref{alg:hw-aware-axc}, Lines 14–16), the total power and area contributions from all CVs are summed, yielding hardware-aware approximate configurations that satisfy the user-defined accuracy threshold while respecting the performance and resource constraints of the selected platform.

\begin{algorithm}[t]
\caption{Hardware-Aware AxC Configurations Generation}
\footnotesize
\label{alg:hw-aware-axc}
\begin{algorithmic}[1]
\REQUIRE Boundary $hawx$ scores per layer; \\ hardware type $HW$ (spatial/temporal); granularity $G$; 
        \\ HW params: Spatial $(\#PEs, SIMD)$, \\   Temporal $(\#CVs \text{, logic/memory resources})$; 
         \\ AxC blocks with base power/area ($p_{base}$, $a_{base}$)
\ENSURE Set of HW-aware AxC configs with est. power/area

\STATE Identify $HW$ type and set params
\STATE $Configs \gets \emptyset$

\FOR{each layer $l$}
    \STATE Select AxC block for $l$ w.r.t. boundary $hawx$
    \IF{$HW =$ Spatial}
        \STATE Map AxC @ granularity $G$ (op/filter/layer) 
        \STATE $cost_{layer} \gets \#\text{AxC in } l \times (p_{base}, a_{base})$
    \ELSE
        \STATE Group layers by $\#CVs$
        \STATE Map AxC @ granularity $G$ (op/filter/layer)
        \STATE $cost_{group} \gets \#\text{AxC in group} \times (p_{base}, a_{base})$
    \ENDIF
\ENDFOR

\STATE $total\_power \gets \sum cost.power_{layer/group}$
\STATE $total\_area \gets \sum cost.area_{layer/group}$
\STATE Append $(AxC\ mapping,\ total\_power,\ total\_area)$ to $Configs$
\RETURN $Configs$
\end{algorithmic}
\end{algorithm}

\section{Experimental Results}
\label{sec:experimental_setup}

\subsection{Experimental Setup}


The proposed framework is evaluated on four representative DNNs: LeNet-5 (2 convolutional and 3 fully connected layers, 97.8\% accuracy ) on MNIST, VGG-11 (8 convolutional and 3 fully connected layers, 92.2\%) and ResNet-18 (20 convolutional and 1 fully connected layer, 93.0\%) on CIFAR-10, and EfficientLiteNet (46 convolutional and 1 fully connected layer, 99.8\%) on the German Traffic Sign Recognition Benchmark (GTSRB). All models are implemented in PyTorch.


To study the impact of approximate arithmetic in this paper, the focus is placed on 8-bit multiplications, as these constitute the dominant operations in DNN workloads. Eight approximate multiplier blocks from the EvoApproxLib library \cite{mrazek2019evoapproxlib} are employed as representative 8-bit AxC primitives, consisting of four from the $K$-series (e.g., 1KV8) and four from the $L$-series (e.g., 1L2N).
 Although multipliers are used in this work, the methodology is general and directly extendable to any approximate adders, multipliers, and higher-precision arithmetic units. 

Software-based evaluations are carried out on a workstation equipped with an Intel Core i7-7500U CPU, 16 GB of RAM, and an NVIDIA GeForce 940MX GPU. To further validate hardware-level behavior, selected approximate configurations for both spatial and temporal DNN accelerators are deployed on an FPGA platform (Xilinx Zynq UltraScale+). Hardware synthesis and implementation are performed using Xilinx Vivado and Vitis HLS 2020.2. As an example of spatial architectures, dataflow-style accelerators are generated by first converting the approximated DNN code into synthesizable C using DeepAxe \cite{taheri2023deepaxe}, followed by hardware synthesis. As an example of temporal architectures, PERUN-style soft processors are employed, in which computations are mapped to several compute vectors on the FPGA fabric. PERUN is a low-power run-time adaptive multi-threaded scalable neural network accelerator designed by the NEUROFABRIX company. Resource allocation and scheduling for both hardware contexts are managed through the HAWX framework.

\subsection{Results and Discussion}
\begin{figure}[t]
    \centering
    \includegraphics[width=0.3\textwidth, height=3.8cm]{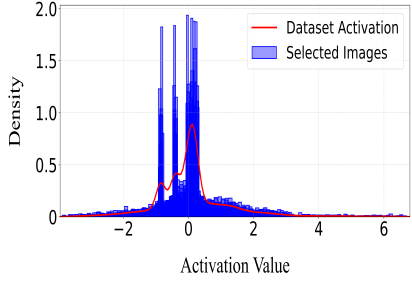}
    \caption{Activation distribution of LeNet-5 on representative inputs}
    \label{fig:hist_lenet}
\end{figure}

Figure \ref{fig:hist_lenet} illustrates the activation histogram of selected input samples during inference for LeNet-5. Due to space constraints, results are shown only for LeNet-5 as a representative example. The red curve corresponds to the activation distribution of the full dataset, while the blue curves represent the distributions of the representative samples selected in Step 1 of HAWX (Figure \ref{fig:hawx_methodology}). The close alignment between the two confirms that the selected samples effectively capture the diversity of the dataset, validating their use for subsequent scoring and exploration.

\begin{figure}[hbtp]
    \centering
    \includegraphics[width=0.3\textwidth, height=3.8cm]{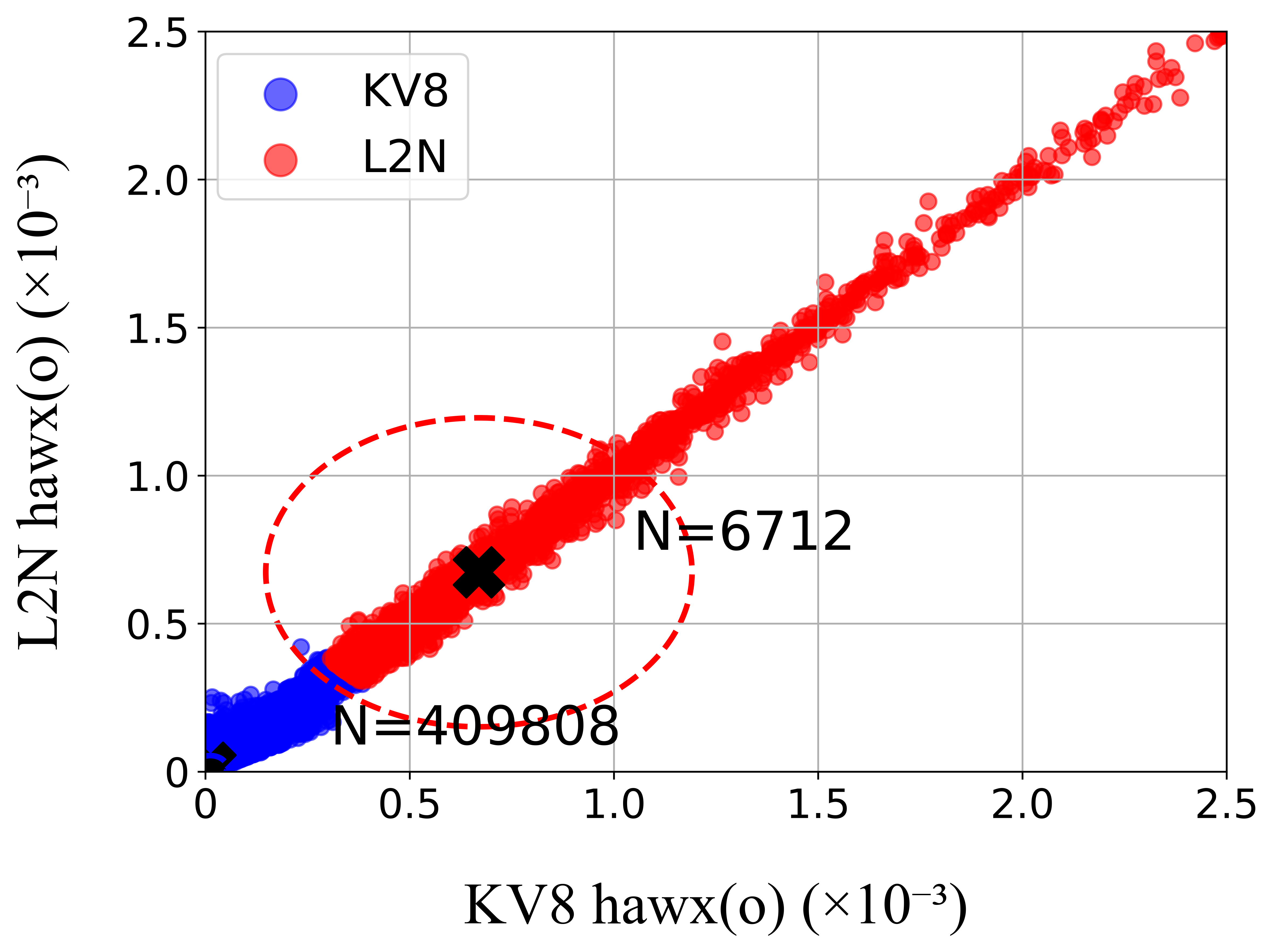}
    \caption{hawx(o) distribution in LeNet-5}
    \label{fig:lenet_hawx}
\end{figure}

Building on this, Figure \ref{fig:lenet_hawx} shows the distribution of $hawx(o)$ scores across all operations in LeNet-5 for two approximate blocks, 1KV8 and 1L2N. The more erroneous 1L2N approximation consistently yields higher $hawx(o)$ values, reflecting greater sensitivity to errors. Table \ref{tab:kv8_l2n_channels_layers} further reports aggregated $hawx(f)$ and $hawx(l)$ values across filters and layers, confirming that applying more aggressive approximations (e.g., 1L2N) induces larger deviations compared to less aggressive ones (e.g., 1KV8).
 These results establish $hawx$ as a fine-grained sensitivity metric that scales naturally from operations to filters and layers.
 
\begin{table}[b]
\vspace{0.2 cm}
\centering
\caption{Comparison of KV8 and L2N $hawx(f)$ and $hawx(l)$ across sample filters and layers.}
\label{tab:kv8_l2n_channels_layers}
\scriptsize
\setlength{\tabcolsep}{3pt} 
\renewcommand{\arraystretch}{1.05} 
\begin{tabular}{|c|c|cc|cc|}
\hline
\textbf{Model} & \textbf{\begin{tabular}[c]{@{}c@{}}AxC\\ Block\end{tabular}}& \multicolumn{2}{c|}{\textbf{Filters}} & \multicolumn{2}{c|}{\textbf{Layers}} \\
\cline{3-6}
 &  & 1 & 2 & 1 & 2 \\
\hline
\multirow{2}{*}{LeNet-5} 
 & KV8 & 6.46E-06 & 1.54E-05 & 1.18E-05 & 1.60E-05 \\
 & L2N & 7.87E-06 & 1.78E-05 & 1.33E-05 & 1.71E-05 \\
\hline
\multirow{2}{*}{VGG-11} 
 & KV8 & 9.16E-05 & 8.53E-05 & 7.90E-05 & 1.30E-06 \\
 & L2N & 9.86E-05 & 9.37E-05 & 8.50E-05 & 1.30E-05 \\
\hline
\multirow{2}{*}{ResNet-18} 
 & KV8 & 3.00E-04& 1.00E-04& 1.81E-04 & 2.43E-04 \\
 & L2N & 2.83E-04 & 1.22E-04 & 1.84E-04 & 2.70E-04 \\
\hline
\multirow{2}{*}{EfficientNet} 
 & KV8 & 1.08E-06 & 1.63E-06 & 7.82E-07 & 1.73E-06 \\
 & L2N & 1.08E-06 & 1.62E-06 & 7.76E-07 & 1.80E-06 \\
\hline
\end{tabular}
\end{table}

\begin{figure*}[ht]
\centering

\begin{subfigure}[b]{0.24\textwidth}
\centering
\begin{tikzpicture}
\begin{axis}[
    width=.9\linewidth,
    height=0.75\linewidth,
    ymin=0, ymax=100,
    ylabel={\scriptsize Accuracy (\%)},
    yticklabel style={rotate=45, font=\scriptsize},
    xtick=data,
    xticklabels={KV6, KV8, KV9, KVP, L2J, L2L, L2N, L12},
    xticklabel style={rotate=45, font=\scriptsize},
    ymajorgrids=true,
    grid style=dashed,
    legend style={font=\scriptsize, legend columns=2}, 
    legend to name=hawxlegend  
]
\addplot[color=blue, mark=o, thick] coordinates {
    (1,97.8) (2,97.78) (3,97.77) (4,97.54) (5,97.71) (6,97.71) (7,95.58) (8,10.28)
};
\addlegendentry{Accuracy};

\addplot[color=red, mark=square, thick] coordinates {
    (1,2.24847e-05*1e6) (2,2.24853e-05*1e6) (3,2.24891e-05*1e6) (4,2.26356e-05*1e6)
    (5,2.26561e-05*1e6) (6,2.30947e-05*1e6) (7,2.40363e-05*1e6) (8,7.05897e-05*1e6)
};
\addlegendentry{HAWX score};
\end{axis}
\end{tikzpicture}
\caption{\footnotesize LeNet-5}
\end{subfigure}
%
\begin{subfigure}[b]{0.24\textwidth}
\centering
\begin{tikzpicture}
\begin{axis}[
    width=.9\linewidth,
    height=0.75\linewidth,
    ymin=0, ymax=100,
    ylabel={\scriptsize Accuracy (\%)},
    yticklabel style={rotate=45, font=\scriptsize},
    xtick=data,
    xticklabels={KV6, KV8, KV9, KVP, L2J, L2L, L2N, L12},
    xticklabel style={rotate=45, font=\scriptsize},
    ymajorgrids=true,
    grid style=dashed
]
\addplot[color=blue, mark=o, thick] coordinates {
    (1,99.7) (2,99.8) (3,92.8) (4,36.4) (5,8.66) (6,0.83) (7,0) (8,0)
};
\addplot[color=red, mark=square, thick] coordinates {
    (1,1.257*16.67) (2,1.457*16.67) (3,1.998*16.67) (4,2.978*16.67)
    (5,2.984*16.67) (6,3.049*16.67) (7,3.189*16.67) (8,5.25*16.67)
};
\end{axis}
\end{tikzpicture}
\caption{\footnotesize EfficientNet}
\end{subfigure}
%
\begin{subfigure}[b]{0.24\textwidth}
\centering
\begin{tikzpicture}
\begin{axis}[
    width=.9\linewidth,
    height=0.75\linewidth,
    ymin=0, ymax=100,
    ylabel={\scriptsize Accuracy (\%)},
    yticklabel style={rotate=45, font=\scriptsize},
    xtick=data,
    xticklabels={KV6, KV8, KV9, KVP, L2J, L2L, L2N, L12},
    xticklabel style={rotate=45, font=\scriptsize},
    ymajorgrids=true,
    grid style=dashed
]
\addplot[color=blue, mark=o, thick] coordinates {
    (1,92.98) (2,92.88) (3,90.96) (4,81.31) (5,89.77) (6,49.37) (7,10) (8,10)
};
\addplot[color=red, mark=square, thick] coordinates {
    (1,926.329/3845.514*100)
    (2,926.484/3845.514*100)
    (3,927.505/3845.514*100)
    (4,955.628/3845.514*100)
    (5,955.351/3845.514*100)
    (6,1027.459/3845.514*100)
    (7,1180.375/3845.514*100)
    (8,3845.514/4200.514*100)
};
\end{axis}
\end{tikzpicture}
\caption{\footnotesize ResNet-18}
\end{subfigure}
%
\begin{subfigure}[b]{0.24\textwidth}
\centering
\begin{tikzpicture}
\begin{axis}[
    width=.9\linewidth,
    height=0.75\linewidth,
    ymin=0, ymax=100,
    ylabel={\scriptsize Accuracy (\%)},
    yticklabel style={rotate=45, font=\scriptsize},
    xtick=data,
    xticklabels={KV6, KV8, KV9, L2J, KVP, L2L, L2N, L12},
    xticklabel style={rotate=45, font=\scriptsize},
    ymajorgrids=true,
    grid style=dashed
]
\addplot[color=blue, mark=o, thick] coordinates {
    (1,92.39) (2,92.26) (3,90.23) (4,90.19) (5,83.18) (6,60.51) (7,11.45) (8,10.15)
};
\addplot[color=red, mark=square, thick] coordinates {
    (1,18.155) (2,20.748) (3,29.061) (4,29.554) (5,37.768) (6,49.825) (7,74.545) (8,78.1426)
};
\end{axis}
\end{tikzpicture}
\caption{\footnotesize VGG-11}
\end{subfigure}


\caption{Comparison of Accuracy (blue) and $hawx(model)$ (red) across DNNs.}
\label{fig:hawx_model_accuracy}
\end{figure*}
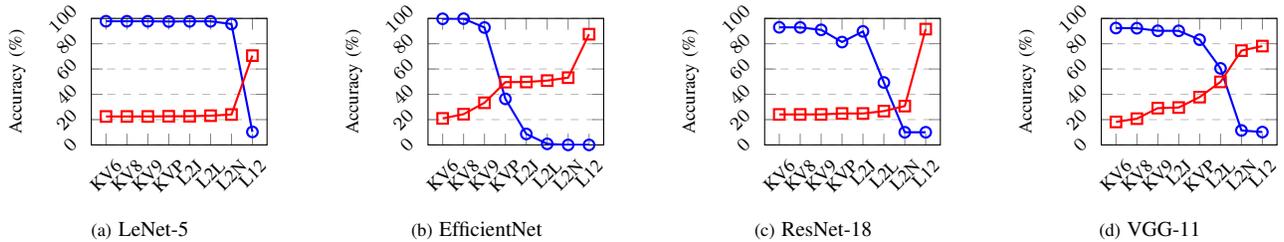

To validate the effectiveness of this metric at the model level, Figure \ref{fig:hawx_model_accuracy} compares the accuracy of models approximated with single AxC blocks to their corresponding $hawx(\text{model})$ scores.  For visibility, $hawx(\text{model})$ values are normalized and mapped to a $0$–$100$ scale. The trends align closely: higher $hawx(\text{model})$ correlates with lower accuracy, while lower scores correspond to minimal degradation. This strong correlation confirms that $hawx$ reliably predicts model-level resilience to approximation.

Representative AxC configurations generated by HAWX are reported in Table \ref{tab:final_result_extended}, together with their accuracy, power, and area metrics. The reported values are normalized for ease of comparison. These configurations are expressed in the format {layers}:{AxC blocks}, where specifying more than one AxC block for a given layer indicates filter-wise approximation. The examples are not exhaustive or necessarily optimal, but rather illustrate different trade-offs across the design space. Importantly, HAWX can propose configurations at multiple granularities, from coarse model-level ({1–47}:KV8) to fine-grained heterogeneous layer- and filter-level (e.g., {1–45,47}:KV8 | {46}:(KV8,KV9)), where specific filters employ different multipliers. Due to space limitations, estimated hardware results are reported only for dataflow accelerators, showing normalized power and area reductions under hardware mapping constraints.

\begin{table}[!b]
\vspace{0.2 cm}
\centering
\scriptsize
\setlength{\tabcolsep}{3pt} 
\renewcommand{\arraystretch}{1.05} 
\caption{Normalized trade-off between accuracy (Acc.), power (Pwr.), and area (Area).}
\label{tab:final_result_extended}
\begin{tabular}{p{1.2cm} p{4.5cm} ccc}
\toprule
\textbf{Model} & \textbf{Configuration} & \textbf{Acc.} & \textbf{Pwr.} & \textbf{Area} \\
\midrule

\multirow{6}{*}{LeNet-5} 
 & \textbf{1-5:KV6} & 97.80 & 1.000 & 1.000 \\
\cmidrule(l){2-5}
 & 1-5:KV9 & 97.77 & 0.965 & 0.939 \\
\cmidrule(l){2-5}
 & 1-2:(KV9,KVP) | 3-5:KV9 & 97.77 & 0.938 & 0.921 \\
\cmidrule(l){2-5}
 & 1-4:KVP | 5:L2J & 97.54 & 0.854 & 0.870 \\
\cmidrule(l){2-5}
 & 1-2:(KVP,L2J) | 3-5:KVP & 97.54 & 0.826 & 0.850 \\
\cmidrule(l){2-5}
 & 1-5:L2N & 95.58 & 0.296 & 0.390 \\
\midrule

\multirow{6}{*}{VGG-11} 
 & \textbf{1-11:KV6} & 92.39 & 1.000 & 1.000 \\
\cmidrule(l){2-5}
 & {1–3,8}:KV6 | {4}:KV8 | {5–7,10–11}:KV9 | {9}:L2N & 92.32 & 0.983 & 0.913 \\
\cmidrule(l){2-5}
 & {1,5,7}:(KV6,KV8) | {2,3,8}:(KV6,KV9) | {4}:KV8 | {6}:KV9 | {9}:L2N | {10,11}:KV9 & 92.32 & 0.859 & 0.798 \\
\cmidrule(l){2-5}
 & {1-5,7}:KV6 | {2,3,8,10}:KV8 | {9}:KV9 | {4,6}:L2J & 92.22 & 0.816 & 0.808 \\
\cmidrule(l){2-5}
 & {1-3}:(KV6,KV8) | {4,10}:KV8 | {5,7}:(KV9,KVP) | {6,9}:KV9 | {8}:{KV8,KV9} | {11}:KV6 & 92.22 & 0.787 & 0.756 \\
\cmidrule(l){2-5}
 & {6,7,9}:KV8 | {4,5}:KV9 | {1-3,10,11}:L2J | {8}:L2N & 88.91 & 0.776 & 0.801 \\
\midrule

\multirow{6}{*}{ResNet-18} 
 & \textbf{1-21:KV6} & 92.39 & 1.000 & 1.000 \\
\cmidrule(l){2-5}
 & 1-21:KV9 & 90.96 & 0.965 & 0.939 \\
\cmidrule(l){2-5}
 & 1-4,6,7,10-20:(KV9,KVP) | 5,8,9,21:KV9 & 90.96 & 0.918 & 0.897 \\
\cmidrule(l){2-5}
 & 3,5,7,10,12,14-18:KVP | 1,2,4,6,8,9,11,13,19,21:L2J & 90.05 & 0.796 & 0.829 \\
\cmidrule(l){2-5}
 & 1-7,10-20:(KVP,L2J) | 8,9,21:KVP & 90.05 & 0.795 & 0.820 \\
\cmidrule(l){2-5}
 & 1,2,4,6,8,9,11,13,19,21:KVP | 3,5,7,10,12,14-18:L2J & 79.95 & 0.771 & 0.811 \\
\midrule

\multirow{6}{*}{EfficientNet} 
 & \textbf{1-47:KV6} & 99.65 & 1.000 & 1.000 \\
\cmidrule(l){2-5}
 & 1-47:KV8 & 99.80 & 0.993 & 0.974 \\
\cmidrule(l){2-5}
 & 1-45,47:KV8 | 46:(KV8,KV9) & 99.80 & 0.993 & 0.972 \\
\cmidrule(l){2-5}
 & 1-47:KV9 & 92.80 & 0.965 & 0.939 \\
\cmidrule(l){2-5}
 & 1-45,47:KV8 | 46:(KV9,L2J) & 92.80 & 0.963 & 0.935 \\
\cmidrule(l){2-5}
 & 1-29,34,42,44-47:KV9 | 30-33,35-41:KVP | 43:L2J & 87.40 & 0.928 & 0.916 \\
\bottomrule
\end{tabular}
\end{table}

\begin{figure}[h]
\centering
\begin{tikzpicture}
\begin{axis}[
    width=0.8\linewidth,          
    height=5cm, 
    view={45}{30},
    xlabel={Accuracy (\%)},
    ylabel={Power},
    zlabel={Area},
    grid=both,
    major grid style={line width=.2pt,draw=gray!50},
    minor tick num=1,
    tick label style={font=\footnotesize},  
    label style={font=\footnotesize},
    legend style={
       draw=none, 
        at={(0.5,-0.25)},            
        anchor=north,
        font=\scriptsize,
        legend columns=4,             
        /tikz/every even column/.append style={column sep=0.3cm}
    },
    legend image post style={scale=0.6}
]

\addplot3[
    only marks,
    mark=square*,
    draw=blue,
    fill=blue!40,
    mark size=2pt
]
coordinates {
    (97.80, 1.201/1.201, 14054/14054)  
    (97.77, 1.197/1.201 , 13910/14054)
};
\addlegendentry{LeNet-5 (PERUN)}

\addplot3[
    only marks,
    mark=square*,
    draw=red,
    fill=red!40,
    mark size=2pt
]
coordinates {
    (92.39, 3.84/3.84 , 101473/101473)  
    (92.22, 3.787/3.84 , 100328/101473) 
};
\addlegendentry{VGG-11 (PERUN)}

\addplot3[
    only marks,
    mark=o,
    draw=blue,
    fill=blue!20,
    mark size=2pt
]
coordinates {
    (97.77, 0.344/0.349 , 36074/36424)
    (97.54, 0.339/0.349, 34144/36424)
    (95.58, 0.337/0.349 , 31707/36424)
    (97.77, 0.343/0.349 , 35672/36424)
    (97.54, 0.337/0.349, 32410/36424) 
};
\addlegendentry{LeNet-5 (Dataflow)}

\addplot3[
    only marks,
    mark=o,
    draw=red,
    fill=red!20,
    mark size=2pt
]
coordinates {
    (92.32, 3.262/3.275 , 245380/247490)
    (92.22 , 3.243/3.275, 243524/247490)
    (88.91 , 3.231/3.275 , 241678/247490)
    (92.32, 3.246/3.275, 243760/247490)
    (92.22, 3.241/3.275 , 243338/247490)
    
};
\addlegendentry{VGG-11 (Dataflow)}

\addplot3[only marks, mark=star, mark options={scale=2, draw=black, fill=yellow}] 
coordinates {(97.80, 1.201/1.201, 14054/14054) }; 

\addplot3[only marks, mark=star, mark options={scale=2, draw=black, fill=yellow}] 
coordinates {(92.39, 3.84/3.84 , 101473/101473)}; 

\end{axis}
\end{tikzpicture}

\caption{Trade-off between accuracy, power, and area for dataflow and PERUN architectures.}
\label{fig:final_result_hls}
\end{figure}
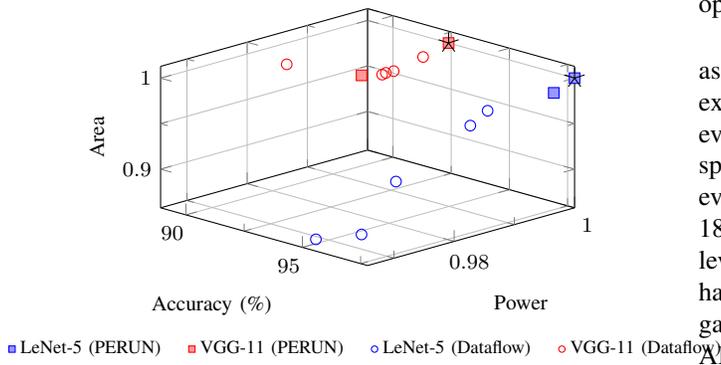

Figure \ref{fig:final_result_hls} further validates these findings, presenting accuracy–power–area trade-offs for LeNet-5 (blue) and VGG-11 (red) on both spatial (dataflow) and temporal (PERUN) architectures. The synthesized results include both layer-wise and filter-wise configurations for both architectures. For the dataflow accelerator, layers are mapped directly onto the hardware, exposing layer-level parallelism (circles). The dataflow configurations correspond to those reported in Table~\ref{tab:final_result_extended}, with the star marker denoting the exact baseline configuration. 

In contrast, temporal execution on the PERUN accelerator (squares) leverages CVs, each comprising multiple PEs with distinct Local IDs (LIDs). Approximation can be selectively applied across LIDs within a CV or across CVs, enabling heterogeneous execution. In these experiments, PERUN configurations employ L2N and KV8 multipliers, with one CV instantiated for LeNet-5 and eight CVs for VGG-11. During execution, the layer or filter is partitioned into work-items; these are grouped into work-groups, which the Work-Group Dispatcher schedules onto available CVs. Each work item is mapped to a specific LID according to the chosen configuration. The results confirm that HAWX-selected AxC configurations achieve substantial reductions in power and area while maintaining accuracy, highlighting gains across both hardware types.

\subsection{Search Runtime Comparison}

The runtime efficiency of HAWX is evaluated against both exhaustive search and state-of-the-art (SOTA) DSE frameworks. Unless otherwise stated, all reported speedups are relative to exhaustive search over the same design space. For LeNet-5, when exploring two EvoApproxLib multipliers (1KV8 and 1L2N), HAWX accelerates layer-level search by $23\times$ and filter-level search by over $10^6\times$. For deeper models such as EfficientLiteNet, filter-level speedups relative to exhaustive search exceed $10^{5000}\times$, demonstrating exponential scalability with network depth and candidate AxC blocks.

GPU-based emulation frameworks such as TFApprox \cite{vaverka2020tfapprox} and I-NN \cite{ahmadilivani2023special} reduce the runtime of a \emph{single configuration evaluation} by $200$–$2800\times$ compared to exact inference, but they still require enumerating the full design space. In contrast, HAWX reduces the number of evaluations from $N_{\text{exh}} = B^{\sum_l f_l}$ (e.g., $10^6$ for LeNet-5, $10^{5000}$ for EfficientLiteNet) down to just a handful values (1–5 evaluations depending on the network), yielding effective end-to-end speedups relative to emulation-based methods.

Predictive metrics such as AME \cite{liu2024architectural} report speedups of up to $10^6\times$ compared to exhaustive evaluation, but only at the model level. They do not support heterogeneous block assignments or hardware-specific constraints. By contrast, HAWX extends predictive sensitivity analysis across multiple abstraction levels (operator, filter, layer, and model) and incorporates hardware costs. This enables speedups on the same baseline of $10^6\times$ for LeNet-5 (filter-level search with 1KV8 and 1L2N) and up to $10^{4000}\times$ for EfficientLiteNet (filter-level search across the same two blocks), while producing hardware-aware Pareto-optimal solutions.

RL-based DSE methods \cite{saeedi2023design} and heuristic approaches such as MAX-DNN \cite{leon2025max} reduce search complexity compared to exhaustive methods, but still require on the order of $10^3$ evaluations. In contrast, HAWX analytically prunes the search space upfront using sensitivity scoring, reducing the number of evaluations to just 1 for LeNet-5, 3 for VGG-11, 4 for ResNet-18, and 5 for EfficientLiteNet. This makes fine-grained, filter-level approximation across heterogeneous AxC blocks and hardware architectures tractable for the first time, bridging the gap between large-scale design exploration and practical edge AI deployment.


\vspace{-0.3cm}
\section{Conclusion}

This work presents \textbf{HAWX}, a hardware-aware scalable exploration framework that employs multi-level sensitivity scoring at different DNN abstraction levels (operator, filter, layer, and model) to guide selective integration of heterogeneous AxC blocks. Supported by predictive models for accuracy, power, and area, HAWX accelerates the evaluation of candidate configurations, achieving over \textbf{23$\times$ speedup} in a layer-level search with two candidate approximate blocks and more than \textbf{(3$\times$10$^{6}$)$\times$ speedup} at the filter-level search only for LeNet-5, while maintaining accuracy comparable to exhaustive search. The HAWX hardware-aware search algorithm supports both spatial and temporal accelerator architectures.
\vspace{-0.3cm}
\section*{Acknowledgements}
\scriptsize
This work was supported in part by the Estonian Research Council grant PUT PRG1467 "CRASHLESS“, EU Grant Project 101160182 “TAICHIP“, by the Deutsche Forschungsgemeinschaft (DFG, German Research Foundation) – Project-ID "458578717", and by the Federal Ministry of Research, Technology and Space of Germany (BMFTR) for supporting Edge-Cloud AI for DIstributed Sensing and COmputing (AI-DISCO) project (Project-ID "16ME1127").

\bibliographystyle{IEEEtran}
\bibliography{ref}

\end{document}